\documentclass{article}

\usepackage{PRIMEarxiv}

\usepackage[utf8]{inputenc} % allow utf-8 input
\usepackage[T1]{fontenc}    % use 8-bit T1 fonts
\usepackage{hyperref}       % hyperlinks
\usepackage{url}            % simple URL typesetting
\usepackage{booktabs}       % professional-quality tables
\usepackage{amsfonts}       % blackboard math symbols
\usepackage{nicefrac}       % compact symbols for 1/2, etc.
\usepackage{microtype}      % microtypography
\usepackage{lipsum}
\usepackage{fancyhdr}       % header
\usepackage{graphicx}       % graphics
\graphicspath{{media/}}     % organize your images and other figures under media/ folder

\usepackage{graphicx} % Required for inserting images
\usepackage{amsmath} 
\usepackage{amsthm} % For theorem-like environments

\usepackage{amssymb} % For \mathbb
\usepackage{natbib}
\usepackage{cleveref} % For \cref and \Cref cross-referencing

 \usepackage{pgfplots}
 \usepackage{pgfplotstable}
 \pgfplotsset{compat=1.18}
 \usepackage{caption}
 \usepackage{subcaption}
 \usepackage{pgfplots}
 \pgfplotsset{compat=1.18}
\pgfplotsset{
  RL4ILbar/.style={
    ybar,
    bar width=15pt,
    width=5cm,
    height=4.0cm,
    ymin=0, ymax=1.12,
    ymajorgrids=true,
    grid style={line width=0.3pt, draw=gray!30},
    axis background/.style={fill=gray!6},
    tick align=inside,
    xtick={1,2},
    xticklabels={No Cam0, No Cam1},
    xticklabel style={font=\scriptsize},
    yticklabel style={font=\scriptsize},
    ytick={0,0.2,0.4,0.6,0.8,1.0},
    enlarge x limits=0.40,
    error bars/y dir=both,
    error bars/y explicit,
    error bars/error bar style={line width=0.6pt, black!60},
    error bars/error mark options={
      rotate=90, black!60, mark size=1.5pt},
  },
  epochline/.style={
    width=0.39\linewidth,
    height=5cm,
    xlabel={\scriptsize Epochs},
    xtick={1,5,10,15,20},
    xticklabels={1,5,10,15,20},
    xmin=0, xmax=21,
    ymin=0, ymax=1.05,
    ytick={0,0.2,0.4,0.6,0.8,1.0},
    tick label style={font=\scriptsize},
    label style={font=\scriptsize},
    title style={font=\small\bfseries},
    axis lines=left,
    grid=major,
    grid style={dotted, gray!40},
  },
  epochline noy/.style={
    epochline,
    yticklabels={,,},
    ylabel={},
  },                          % <-- comma here
  RL4ILtopk/.style={
    ybar=2pt,
    bar width=6.0pt,
    width=1.2\linewidth,
    height=4.0cm,
    ymin=0, ymax=1.12,
    ymajorgrids=true,
    grid style={line width=0.3pt, draw=gray!30},
    axis background/.style={fill=gray!6},
    tick align=inside,
    xtick={1,2,3,4,5},
    xticklabels={4, 8, 12, 16, 32},
    xlabel={\scriptsize $K$},
    xticklabel style={font=\scriptsize},
    yticklabel style={font=\scriptsize},
    ytick={0,0.2,0.4,0.6,0.8,1.0},
    enlarge x limits=0.12,
    error bars/y dir=both,
    error bars/y explicit,
    error bars/error bar style={line width=0.6pt, black!60},
    error bars/error mark options={rotate=90, black!60, mark size=1.5pt},
  }
}
 \usepackage{subcaption}
 \usepackage{booktabs}
 \usepackage{multirow}
 \usepackage{xcolor}

\title{Reinforcement Learning-Guided Retrieval with Soft Fusion for Robust Multimodal Imitation Learning under Missing Modalities
}

\author{
  Hassan Ismkhan* and Hamid Bouchahcia \\
  Bournemouth University \\
  Bournemouth, UK \\
  \{hismkhan*, abouchachia\}@bournemouth.ac.uk \\ \\
  *Responsible author
}

\begin{document}
\begin{large}  \begin{center}In the name of God\end{center} 
\end{large} 

\vspace{2\baselineskip}

\maketitle

%===============================================================================

\begin{abstract}
     Robotic systems perceive the world through multiple input modalities --- including visual camera streams and natural language instructions --- and must select appropriate actions based on these signals. However, assuming the permanent availability of all input devices is unrealistic, as sensors may fail, become occluded, or drop out entirely during deployment. Robust handling of such missing-modality scenarios is therefore essential for real-world robot operation. This paper introduces RL4IL, a reinforcement learning guided method for imitation learning that selects the most suitable action for a given observation by identifying the most relevant expert demonstrations from a training library. A reinforcement learning policy, trained via Proximal Policy Optimisation over Breadth-First Search candidate sets, ranks candidate demonstrations and a soft cross-attention fusion head aggregates their action signals to produce the final prediction. When a modality is missing at inference time, a dedicated per-modality RL retrieval policy identifies donor demonstrations from the training library, and a soft imputation head reconstructs the missing embedding via cross-attention over the top-ranked donors --- without requiring any retraining of the system. Experiments on three LIBERO benchmark suites demonstrate that RL4IL substantially outperforms state-of-the-art imitation learning methods under sensor dropout conditions, while requiring no policy network training. The code can be found at \url{https://github.com/h-ismkhan/Reinforcement-Learning-via-kNN-for-Robotic-Learning-with-Missing-Camera}
\end{abstract}


\section{Introduction}
\label{sec:introduction}
Robotic manipulation systems deployed in real-world environments must process heterogeneous sensory streams — including visual observations from multiple camera viewpoints and natural-language task instructions \cite{brohan2024rt, liu2023libero}. While recent advances in imitation learning have demonstrated impressive performance on structured benchmarks \cite{mandlekar2021robomimic, reuss2023beso, Zhao-RSS-23, chi2025diffusion}, these methods operate under a tacit assumption that all sensor modalities remain available and intact throughout deployment. In practice, however, cameras can fail due to hardware faults, become occluded by objects or the robot's own body, or suffer from signal degradation in uncontrolled settings \cite{peri2024masked}. Robust robot behaviour under such sensor dropout conditions is therefore not merely a desirable property but a fundamental requirement for real-world deployment — yet it remains largely unaddressed in the imitation learning literature, where even methods explicitly designed for robustness \cite{vanjani2025disdp} show limited recovery under complete camera failure.

Although significant progress has been made in multimodal learning
with missing modalities~\citep{li2025simmlm,zhi2025borrowing,
zhao2025enhancing,lee2023multimodal,cheng2026vamp}, these methods
are not directly applicable to robotic manipulation for three
compounding reasons.
First, several are confined to homogeneous, domain-specific modalities:
MRI sequences~\citep{liu2023m3ae,wang2023multi} or physiological
signals~\citep{shen2024robust}, rather than the heterogeneous
vision-and-language streams of robot learning.
Second, methods that do span broader modality types --- including
omni-modal retrieval~\citep{zuo2026reid5o}, visual
recognition~\citep{lee2023multimodal,cheng2026vamp}, and general
classification or segmentation~\citep{li2025simmlm,zhi2025borrowing,
zhao2025enhancing} --- target static prediction tasks and cannot
produce the sequential action outputs required for manipulation.
Third, and most critically for deployment, \emph{all} of these
approaches require retraining or fine-tuning of learnable parameters
to accommodate a new missing-modality pattern, making zero-shot sensor
dropout at inference time infeasible.

\noindent\textbf{Summary of contributions.}
\begin{itemize}
    \item \textbf{RL-guided demonstration retrieval.}
    We introduce the first use of reinforcement learning to select the
    most relevant expert demonstration from a training library for
    imitation learning.
    A PPO-trained policy operates over BFS-augmented candidate sets and
    learns to rank neighbours by their usefulness to the current query,
    replacing heuristic distance-weighted aggregation with a learned
    selection strategy.

    \item \textbf{Soft cross-attention fusion.}
    Rather than committing to a single retrieved demonstration, a
    lightweight cross-attention fusion head performs a soft, attended
    aggregation over the top-$K'$ RL-ranked candidates, producing more
    reliable action predictions than hard argmax selection --- particularly
    under noisy retrieval conditions.

    \item \textbf{Zero-shot missing-modality robustness without retraining.}
    RL4IL requires no policy retraining to handle sensor dropout at
    inference time.
    When a camera fails or becomes occluded, a dedicated per-modality RL
    retrieval policy identifies donor demonstrations from a frozen training
    library, and a soft imputation head reconstructs the missing embedding
    via cross-attention over the top-ranked donors --- leaving all other
    pipeline components unchanged.

    \item \textbf{State-of-the-art performance on LIBERO benchmarks.}
    Experiments across LIBERO-Spatial, LIBERO-Object, and LIBERO-Goal
    demonstrate that RL4IL substantially outperforms all baselines ---
    including BC, DisBC, BESO-ACT, and DisDP --- under complete camera
    dropout, while requiring no policy network training.
\end{itemize}
The remainder of this paper is organised as follows:
Section~\ref{sec:related} reviews related work, Section~\ref{sec:method}
details the RL4IL framework, Section~\ref{sec:experiments} presents
experimental results and ablation studies, and Section~\ref{sec:conclusion}
concludes.
%===============================================================================
\section{Related Work}
\label{sec:related}
\textbf{Imitation learning for manipulation.}
Behaviour Cloning~\citep{mandlekar2021robomimic} and its descendants ---
including diffusion policies~\citep{reuss2023beso}, action-chunking
transformers~\citep{Zhao-RSS-23}, and a growing family of
Vision-Language-Action models~\citep{zhao2025cot,hou2025dita,li2026cogvla,
zhang20264d,huang2026thinkact,wang2025vq,dreamvla25,wang2026vla,
tian2025predictive} --- have achieved strong results on structured
benchmarks such as LIBERO~\citep{liu2023libero}.
However, all of these methods assume every training modality remains
available at deployment.
Camera position has been identified as the largest single source of
generalisation failure in visual IL~\citep{xie2024decomposing}, and
demonstration quality is a key determinant of policy
performance~\citep{li2025train,mandlekar2021robomimic}.
 
\textbf{Robustness to sensor dropout.}
Masked-modality training for RL policies~\citep{peri2024masked} and
multi-view disentanglement~\citep{dunion2024multiview,becker2024combining}
improve robustness to camera failure in RL settings.
On the IL side, DisDP~\citep{vanjani2025disdp} structures sensor inputs
into shared and private representations to reduce reliance on individual
cameras, but still requires full policy retraining and collapses under
\emph{complete} camera dropout (Table~\ref{table:dropout-comparison}).
 
\textbf{Retrieval-based policies.}
Recent retrieval-augmented IL methods condition policies on retrieved
demonstrations~\citep{reuss2023beso,Zhao-RSS-23}, but all assume complete
observations at query time.

\textbf{Multimodal Learning.}
Missing-modality multimodal learning has attracted considerable
attention, but predominantly in domains and settings orthogonal to
ours.
Domain-specific methods --- M$^3$AE~\citep{liu2023m3ae} and
ShaSpec~\citep{wang2023multi} for missing MRI sequences,
CIMSleepNet~\citep{shen2024robust} for incomplete physiological
signals --- require full retraining and do not generalise beyond
their target modalities.
More general frameworks such as SimMLM~\citep{li2025simmlm} and the
in-context retrieval approach in~\citep{zhi2025borrowing} span
broader task types but are confined to static classification or
segmentation and rely on end-to-end trainable architectures.
Prompt-based~\citep{lee2023multimodal,zhao2025enhancing,cheng2026vamp}
and omni-modal~\citep{zuo2026reid5o} methods target visual
recognition or retrieval and still require updating learnable
parameters to handle new missing-modality patterns.

\section{Method}
\label{sec:method}

   We consider a robot that learns from a library of expert demonstrations
    $\mathcal{D}_{\mathrm{tr}} = \{(\mathbf{x}_i, y_i)\}_{i=1}^{N}$, where
    each demonstration $\mathbf{x}_i$ is observed through $M$ sensor modalities
    --- such as RGB cameras and natural-language task instructions --- and
    $y_i$ denotes the associated supervision signal, which is a recorded
    action sequence in imitation learning settings or a scalar label in
    regression and classification tasks.
    At deployment, one or more sensors may fail or become unavailable,
    producing a test observation $\mathbf{x}$ with a \emph{missing-modality
    subset}.
    Rather than retraining the system or relying on hand-crafted imputation
    rules, our method retrieves the most relevant expert demonstration from
    $\mathcal{D}_{\mathrm{tr}}$ and replays its recorded response, adapting
    gracefully to sensor dropout through a learned imputation and retrieval
    pipeline.
 
    For each modality $m \in \{1,\dots,M\}$, a frozen pretrained encoder
    $E_m : \mathcal{X}_m \to \mathbb{R}^{d_m}$ maps the raw sensory input
    to a $d_m$-dimensional embedding
    $\mathbf{z}_i^{(m)} = E_m\!\left(\mathbf{x}_i^{(m)}\right)$.
    No encoder weights are updated during training; the robot's perceptual
    representations are fixed, and all learning occurs in the retrieval
    and imputation modules described in this section. 
    
    To place all modalities on a common footing, we apply modality-specific
    $z$-score normalisation to each embedding block and define a
    modality-fair distance that weights each modality's contribution equally
    regardless of its embedding dimensionality; full details are given in
    Appendix~\ref{app:norm-dist}.
% -----------------------------------------------------------
\subsection{Neighbours Sets}
\label{sec:method:sets}
   For a query observation $\mathbf{o}$, the candidate neighbour set
    $\mathcal{C}(\mathbf{o})$ is defined as the $k$ training demonstrations
    closest to $\mathbf{o}$ under the modality-fair distance $d$: 
    \begin{equation}
        \mathcal{C}_{\mathrm{kNN}}(\mathbf{o})
            = \operatorname{arg\,top\text{-}}k_{\mathbf{n}\in\mathcal{D}_{\mathrm{tr}}}
              \bigl[-d(\mathbf{o},\mathbf{n})\bigr].
        \label{eq:knn-set}
    \end{equation} 
    \noindent The neighbourhood size $k$ is a fixed hyperparameter shared
    across all experiments.
    These $k$ demonstrations form the seed set from which the BFS-augmented
    candidate pool is constructed in the following subsection.
\subsection{BFS-Based Candidate Set and RL Selection}
\label{sec:method:rl}
    The base prediction aggregates all members of $\mathcal{C}(\mathbf{o})$
    equally (subject to distance weighting), but not all neighbours are equally
    informative.
    We therefore train a reinforcement learning (RL) policy that selects a
    \emph{single} best neighbour from $\mathcal{C}(\mathbf{o})$, replacing
    the weighted aggregate with a learned point prediction.
\subsubsection{BFS-Augmented Candidate Set}
\label{sec:method:bfs}
    To give the RL agent a richer and more label-diverse pool to choose from,
    we augment each training point's initial neighbour set through a
    breadth-first search (BFS) on a training $k$-NN graph.    
    Formally, let $G = (V, E)$ be the directed graph whose vertex set is
    $V = \mathcal{D}_{\mathrm{tr}}$ and whose edge set is
    \begin{equation}
        E = \bigl\{(\mathbf{s},\mathbf{v},w_{\mathbf{sv}})\;\big|\;
            \mathbf{v}\in\mathcal{C}(\mathbf{s}),\;
            w_{\mathbf{sv}} = d(\mathbf{s},\mathbf{v})\bigr\},
        \label{eq:graph}
    \end{equation}
    where $\mathcal{C}(\mathbf{s})$ is the initial neighbour set of $\mathbf{s}$
    ($k$-NN).    
    For a training source $\mathbf{s}$ with label $y_{\mathbf{s}}$, we run a
    depth-limited shortest-path search (Dijkstra with depth bound $D$) from
    $\mathbf{s}$ outward through $G$.
    For every visited node $\mathbf{v}$, the graph distance
    $\delta(\mathbf{s},\mathbf{v})$ is the length of the shortest path from
    $\mathbf{s}$ to $\mathbf{v}$ using at most $D$ hops; this distance is updated
    during the search whenever a shorter path is found, consistent with
    correct Dijkstra behaviour.
    The BFS candidate set of $\mathbf{s}$ is then    
    \begin{equation}
        \mathcal{B}(\mathbf{s})
            = \bigl\{\mathbf{v}\in V \setminus\{\mathbf{s}\}\;\big|\;
              \delta(\mathbf{s},\mathbf{v})<\infty,\;
              \text{depth}(\mathbf{s},\mathbf{v})\leq D\bigr\},
        \label{eq:bfs-set}
    \end{equation}    
    \noindent together with their graph distances
    $\{\delta(\mathbf{s},\mathbf{v})\}_{\mathbf{v}\in\mathcal{B}(\mathbf{s})}$.
    We retain $\mathcal{B}(\mathbf{s})$ only if it contains at least one node
    whose label matches $y_{\mathbf{s}}$; otherwise the point is excluded from
    RL training.    
    The \emph{oracle} for source $\mathbf{s}$ is the node in
    $\mathcal{B}(\mathbf{s})$ with the same label and the smallest graph
    distance:    
    \begin{equation}
        \mathbf{n}^*(\mathbf{s})
            = \operatorname*{arg\,min}_{\mathbf{v}\in\mathcal{B}(\mathbf{s}),\;
                                        y_{\mathbf{v}}=y_{\mathbf{s}}}
              \delta(\mathbf{s},\mathbf{v}).
        \label{eq:oracle}
    \end{equation}    
    \noindent Before presenting $\mathcal{B}(\mathbf{s})$ to the policy, its
    members are \emph{randomly shuffled} at each training epoch, so the oracle
    is not always at a predictable position and the policy cannot exploit
    ordering artefacts.
\subsubsection{Policy}
\label{sec:method:policy}
    The policy is parameterised by an attention-style scoring head that assigns a
    scalar score to every candidate in $\mathcal{B}(\mathbf{s})$.
    Two encoders first produce fixed-size representations:    
    \begin{align}
        \mathbf{h}_{\mathbf{s}}  &= \mathrm{MLP}_q(\phi(\mathbf{s})),
        \label{eq:h-query}\\
        \mathbf{h}_{\mathbf{v}}  &= \mathrm{MLP}_c(\psi(\mathbf{s},\mathbf{v})),
        \quad \mathbf{v}\in\mathcal{B}(\mathbf{s}),
        \label{eq:h-cand}
    \end{align}    
    \noindent where $\phi(\mathbf{s})$ is the state feature vector for the source
    and $\psi(\mathbf{s},\mathbf{v})$ is the candidate feature vector.
    Their definitions are given in Appendix~\ref{app:method:features}.
    The score for candidate $\mathbf{v}$ is    
    \begin{equation}
        \mathrm{score}(\mathbf{v})
            = \mathrm{MLP}_\theta\!\left(
              \bigl[\mathbf{h}_{\mathbf{s}};\,
                    \mathbf{h}_{\mathbf{v}};\,
                    \mathbf{h}_{\mathbf{s}}\odot\mathbf{h}_{\mathbf{v}}\bigr]
              \right),
        \label{eq:score}
    \end{equation}    
    \noindent where $[\,\cdot\,;\,\cdot\,]$ denotes concatenation and $\odot$
    element-wise multiplication.
    The policy is defined via a softmax over all $K=|\mathcal{B}(\mathbf{s})|$
    candidates:    
    \begin{equation}
        \pi_\theta(a=i \mid \mathbf{s})
            = \frac{\exp\!\bigl(\mathrm{score}(\mathbf{v}_i)\bigr)}
                   {\displaystyle\sum_{j=1}^{K}
                    \exp\!\bigl(\mathrm{score}(\mathbf{v}_j)\bigr)},
        \label{eq:policy}
    \end{equation}    
    \noindent and the predicted label is $\hat{y}(\mathbf{s}) =
    y_{\mathbf{n}_{\hat{a}}}$, where $\hat{a}=\operatorname{arg\,max}_i
    \pi_\theta(a=i\mid\mathbf{s})$ at test time.
\subsubsection{Reward and Training Objective}
\label{sec:method:reward}
    The immediate reward for selecting action $a$ (i.e.\ candidate
    $\mathbf{v}_a$) is defined relative to the oracle:    
    \begin{equation}
        r(\mathbf{s},a)
            = -\mathcal{L}\!\bigl(y_{\mathbf{v}_a},\,y_{\mathbf{s}}\bigr)
              +\mathcal{L}\!\bigl(y_{\mathbf{n}^*(\mathbf{s})},\,y_{\mathbf{s}}\bigr),
        \label{eq:reward}
    \end{equation}    
    \noindent where $\mathcal{L}$ is the task-specific loss and
    $\mathbf{n}^*(\mathbf{s})$ is the oracle defined in~\eqref{eq:oracle}.
    For classification tasks we use the $0/1$ loss, giving
    $r(\mathbf{s},a) = \mathbf{1}[y_{\mathbf{v}_a}=y_{\mathbf{s}}]
                      -\mathbf{1}[y_{\mathbf{n}^*(\mathbf{s})}=y_{\mathbf{s}}]$.
    Because the oracle always achieves the minimum possible loss within the
    candidate set, $r\in\{-1,0\}$: the reward is $0$ when the policy matches
    the oracle and $-1$ when it selects a wrong-label candidate.    
    The policy is trained with Proximal Policy Optimisation (PPO) using the
    clipped surrogate objective~\citep{schulman2017proximal}:    
    \begin{equation}
        \mathcal{J}_{\mathrm{PPO}}(\theta)
            = \mathbb{E}\!\left[
              \min\!\left(
                \rho_t(\theta)\,\hat{A}_t,\;
                \operatorname{clip}\!\left(\rho_t(\theta),\,
                                          1{-}\varepsilon,\,
                                          1{+}\varepsilon\right)\hat{A}_t
              \right)
              \right]
              + \alpha_{\mathrm{ent}}\,\mathcal{H}(\pi_\theta),
        \label{eq:ppo}
    \end{equation}    
    \noindent where $\rho_t(\theta)=\pi_\theta(a_t\mid\mathbf{s}_t)/
    \pi_{\theta_{\mathrm{old}}}(a_t\mid\mathbf{s}_t)$ is the probability
    ratio, $\hat{A}_t=r(\mathbf{s},a_t)$ is the single-step advantage (no
    value baseline is used), $\varepsilon$ is the clipping threshold,
    $\mathcal{H}(\pi_\theta)$ is the entropy of the policy distribution, and
    $\alpha_{\mathrm{ent}}$ is the entropy bonus coefficient.
    Only training points with $|\mathcal{B}(\mathbf{s})|\geq 2$ are used,
    since a singleton set admits no meaningful choice.
    
    After each epoch the policy is evaluated on a held-out validation split and
    the checkpoint with the best validation $F_1$-micro is retained for final
    test evaluation.

\subsubsection{Soft Candidate Fusion via Cross-Attention}
    Hard argmax selection of a single neighbour discards potentially useful
    signal from the remaining high-ranked candidates in $\mathcal{B}(\mathbf{s})$.
    To address this, we augment the pipeline with a lightweight \emph{fusion
    head} that performs a soft, attended aggregation over the top-$K'$ candidates
    ranked by the RL policy.
    After the policy assigns a score $r_j$ to every node in the candidate set,
    the $K'$ highest-scoring candidates are retained.
    Let $\mathbf{q} \in \mathbb{R}^{D}$ denote the query embedding and
    $\{\mathbf{v}_j\}_{j=1}^{K'} \subset \mathbb{R}^{D}$ the selected candidate
    embeddings with associated labels $\{y_j\}_{j=1}^{K'}$.
    The fusion head projects query and candidates into a shared $d_f$-dimensional
    space using separate linear layers, applies $H$-head scaled dot-product
    attention to obtain per-candidate weights $\{\alpha_j\}$, and computes a
    soft attended label $\tilde{y} = \sum_j \alpha_j y_j$.
    A two-layer MLP then refines $\tilde{y}$ using the attention-weighted context
    vector $\mathbf{c} = \sum_j \alpha_j \mathbf{v}_j^\prime$ (where
    $\mathbf{v}_j^\prime$ denotes the projected key), producing the final scalar
    prediction $\hat{y} = \mathrm{MLP}([\mathbf{c};\,\tilde{y}])$.
    The same fusion mechanism extends straightforwardly to classification tasks.
    In that setting, each candidate label $y_j \in \{0,\dots,C-1\}$ is first
    mapped to a one-hot vector $\mathbf{e}_j \in \{0,1\}^C$, and the soft
    attended representation becomes $\tilde{\mathbf{p}} = \sum_j \alpha_j
    \mathbf{e}_j \in \Delta^{C-1}$, which can be interpreted as a
    label-distribution estimate over the neighbourhood.
    The MLP head then takes $[\mathbf{c};\,\tilde{\mathbf{p}}] \in
    \mathbb{R}^{d_f + C}$ as input and produces logits over $C$ classes,
    trained with cross-entropy loss.
    At inference, the predicted class is $\hat{y} =
    \arg\max_c \mathrm{MLP}([\mathbf{c};\,\tilde{\mathbf{p}}])_c$.
    The remainder of the architecture — query/key projections, multi-head
    attention, and context vector computation — is identical to the regression
    case, so a single implementation supports both tasks by switching only the
    output layer and loss function.

\subsubsection{Missing-Modality Imputation via Per-Modality RL.}
    When a camera fails or becomes occluded during robot operation, the
    corresponding modality block is absent from the query embedding $\mathbf{q}$,
    rendering the concatenated representation incomplete and causing the
    modality-fair distance to an unreliable basis for retrieval.
    We address this with a two-stage imputation pipeline that precedes label
    prediction.
    For each potentially missing modality $m$,
    a dedicated RL policy $\pi^{(m)}_\phi$ is trained to select the best
    \emph{donor} from a modality-$m$-restricted kNN index of the training set
    -- only training samples where modality $m$ is available are eligible
    donors.
    The partial query embedding $\tilde{\mathbf{q}}$ is formed from the
    remaining $M' < M$ present modalities using re-normalised modality-fair
    scaling with $M'$ in place of $M$.
    A BFS candidate set is built over this donor index, and the imputation
    oracle is defined as the donor whose modality-$m$ embedding
    $\mathbf{z}_j^{(m)}$ minimises the squared $\ell_2$ distance to the
    ground-truth embedding $\mathbf{z}_i^{(m)}$.
    The policy is optimised with PPO using the same ranking reward
    $r = (\mathrm{rank}_{\mathrm{oracle}} - \mathrm{rank}_{\mathrm{action}})
    / (K{-}1) \in [-1,+1]$, where candidates are ranked by their squared
    $\ell_2$ distance to the ground truth.
    At training time the ground-truth modality-$m$ embedding is available and
    used for oracle definition and BFS seeding; at inference it is replaced by
    a zero-vector cold start, and the policy selects donors based on the
    partial-query context alone.
     
    \paragraph{Soft Imputation via Attended Donor Aggregation.}
    Hard selection of a single donor copies one embedding vector verbatim,
    which can be noisy if no single donor closely matches the missing
    modality.
    To address this, we optionally augment the imputation stage with a
    \emph{soft imputation head} that aggregates the top-$K_{\mathrm{imp}}'$ RL-ranked donors via cross-attention, analogous to the prediction fusion head.
    Let $\{\mathbf{d}_j^{(m)}\}_{j=1}^{K_{\mathrm{imp}}'}$ denote the
    selected donor embeddings.
    The head projects the partial query $\tilde{\mathbf{q}}$ and each donor
    into a shared $d_{\mathrm{imp}}$-dimensional space via separate linear
    layers, applies $H_{\mathrm{imp}}$-head scaled dot-product attention to
    obtain weights $\{\beta_j\}$, and computes a soft attended embedding
    $\tilde{\mathbf{z}}^{(m)} = \sum_j \beta_j \mathbf{d}_j^{(m)}$.
    A two-layer MLP then refines $\tilde{\mathbf{z}}^{(m)}$ using the
    attention-weighted key context, producing the final imputed block
    $\hat{\mathbf{z}}^{(m)} = \mathrm{MLP}([\tilde{\mathbf{z}}^{(m)};\,
    \mathbf{c}^{(m)}]) \in \mathbb{R}^{d_m}$.
    The soft imputation head is trained with a supervised MSE objective
    $\mathcal{L} = \|\hat{\mathbf{z}}^{(m)} - \mathbf{z}_i^{(m)}\|_2^2$
    on training samples where modality $m$ is present (simulated as missing),
    with the imputation RL policy frozen throughout.
    The resulting imputed block replaces the missing entry in the raw embedding,
    after which the full embedding is reconstructed with $M{=}3$ modalities
    and passed to the prediction pipeline.

\subsubsection{Extension to Imitation Learning}
\label{sec:method:il}
 
    Although the preceding formulation is stated for supervised prediction
    tasks, the same pipeline applies directly to imitation learning (IL)
    settings in which each training sample is an expert demonstration rather
    than a labelled input--output pair.
    In this context, a demonstration consists of a sequence of multimodal
    observations --- such as visual frames from one or more cameras and a
    natural-language task instruction --- together with the corresponding
    recorded action sequence.
    Each demonstration is treated as a single multimodal sample
    $\mathbf{x}_i$: frozen pretrained encoders map each observation modality
    to a fixed-dimensional embedding, the embeddings are modality-fair
    normalised and concatenated to form $\mathbf{f}_i$, and the action
    sequence plays the role of the ``label'' $y_i$.
    At test time, the initial observation of a new episode is embedded in
    the same way, the RL policy with fusion head selects the most relevant
    training demonstration from the BFS candidate set, and the corresponding
    action sequence is replayed open-loop in the environment.
    Task success thus depends entirely on the quality of the retrieved
    demonstration, requiring no policy network training.
    When one or more observation modalities are unavailable at test time ---
    for example due to camera failure --- the missing-modality imputation
    pipeline (Section~\ref{sec:method:rl}) reconstructs the absent embedding
    block from donor demonstrations before retrieval proceeds, leaving the
    rest of the pipeline unchanged.

% -----------------------------------------------------------

% ================================================================
%  IN-TEXT EXPLANATION — paste before Figure 1 in your results section
% ================================================================
\section{Experiments}
\label{sec:experiments}
 
In this section we report results of RL4IL on three robotic manipulation
benchmarks from the LIBERO suite~\cite{liu2023libero}: LIBERO-Spatial,
LIBERO-Object, and LIBERO-Goal.
LIBERO-Spatial consists of ten tasks that require spatial reasoning about
object positions on a tabletop; LIBERO-Object consists of ten tasks
centred on object-centric manipulation under varying object configurations;
and LIBERO-Goal consists of ten tasks defined by goal-conditioned
instructions that require longer-horizon reasoning.
All three suites share the same observation space: an agent-view RGB
camera, an in-hand RGB camera, and a natural-language task instruction,
giving $M{=}3$ modalities per demonstration.
Each suite provides 50 expert demonstrations per task recorded via
teleoperation.
 
For all three suites we evaluate under the two conditions: (i)~agent-view camera absent (cam\,0 masked), and
(ii)~in-hand camera absent (cam\,1 masked).
In these conditions, the missing camera embedding is reconstructed
by our per-modality imputation pipeline before retrieval proceeds; 
Performance is measured as task success rate: the fraction of rollouts that complete the task within 260 steps, averaged over 3 random seeds with 25 rollouts per task per seed.
We compare against these baselines:
Behaviour Cloning~(BC), Disentangled BC~(DisBC), BESO-ACT, BESO-ACT with
modality dropout training~(BESO-ACT-Dropout), and the proposed
Disentangled Diffusion Policy~(DisDP).
Results are shown in \cref{table:dropout-comparison}. 

% ================================================================
%  Add to preamble if not already present:
%  \usepackage{booktabs}
%  \usepackage{multirow}
%  \usepackage{bm}
%
%  No custom \best command needed — we use \mathbf{} directly.
% ================================================================
\begin{table*}[t]
\caption{
  Success rate under complete camera dropout
}
\label{table:dropout-comparison}
\setlength{\tabcolsep}{5pt}
\centering
\makebox[\textwidth][c]{%
\begin{tabular}{llcccccc}
\toprule
\textbf{Suite} & \textbf{Config} &
  \textbf{BC} & \textbf{DisBC} & \textbf{BESO-ACT} &
  \textbf{BESO-ACT-DO} & \textbf{DisDP} & \textbf{RL4IL(Ours)} \\
\midrule
 
\multirow{2}{*}{LIBERO-Goal}
  & No Cam0
    & $0.000 \pm 0.00$
    & $0.000 \pm 0.00$
    & $0.084 \pm 0.01$
    & $0.040 \pm 0.00$
    & $0.004 \pm 0.00$
    & $\mathbf{0.700 \pm 0.227}$ \\[2pt]
  & No Cam1
    & $0.000 \pm 0.00$
    & $0.000 \pm 0.00$
    & $0.012 \pm 0.00$
    & $0.004 \pm 0.00$
    & $0.200 \pm 0.04$
    & $\mathbf{0.705 \pm 0.185}$ \\
 
\midrule
 
\multirow{2}{*}{LIBERO-Object}
  & No Cam0
    & $0.000 \pm 0.00$
    & $0.110 \pm 0.03$
    & $0.204 \pm 0.00$
    & $0.004 \pm 0.00$
    & $0.295 \pm 0.04$
    & $\mathbf{0.733 \pm 0.262}$ \\[2pt]
  & No Cam1
    & $0.000 \pm 0.00$
    & $0.000 \pm 0.00$
    & $0.012 \pm 0.01$
    & $0.000 \pm 0.00$
    & $0.226 \pm 0.03$
    & $\mathbf{0.677 \pm 0.216}$ \\
 
\midrule
 
\multirow{2}{*}{LIBERO-Spatial}
  & No Cam0
    & $0.000 \pm 0.00$
    & $0.000 \pm 0.00$
    & $0.028 \pm 0.00$
    & $0.023 \pm 0.00$
    & $0.144 \pm 0.02$
    & $\mathbf{0.540 \pm 0.200}$ \\[2pt]
  & No Cam1
    & $0.000 \pm 0.00$
    & $0.004 \pm 0.00$
    & $0.004 \pm 0.00$
    & $0.023 \pm 0.04$
    & $0.112 \pm 0.00$
    & $\mathbf{0.441 \pm 0.198}$ \\
 
\bottomrule
\end{tabular}%
}
\end{table*}

\begin{figure}[t]
\centering
\begin{center}
\begin{tikzpicture}
  \draw[fill=blue!65, draw=blue!80] (0,0) rectangle (0.25,0.45);
  \node[anchor=west, font=\scriptsize] at (0.32,0.22) {With Soft Fusion};
  \draw[fill=red!65, draw=red!80] (3.2,0) rectangle (3.45,0.45);
  \node[anchor=west, font=\scriptsize] at (3.52,0.22) {Without Soft Fusion};
\end{tikzpicture}
\end{center}
\vspace{2pt}
\begin{subfigure}[b]{0.32\linewidth}
\centering
\begin{tikzpicture}
\begin{axis}[RL4ILbar,
  title={\small\textbf{LIBERO-Goal}},
  ylabel={\scriptsize Success Rate},
  ylabel style={yshift=-2pt}]
\addplot[fill=blue!65,draw=blue!80,error bars/.cd,y dir=both,y explicit]
  coordinates{(1,0.700)+-(0,0.227)(2,0.705)+-(0,0.185)};
\addplot[fill=red!65,draw=red!80,error bars/.cd,y dir=both,y explicit]
  coordinates{(1,0.651)+-(0,0.265)(2,0.729)+-(0,0.207)};
\end{axis}
\end{tikzpicture}
\end{subfigure}
\begin{subfigure}[b]{0.32\linewidth}
\centering
\begin{tikzpicture}
\begin{axis}[RL4ILbar,
  title={\small\textbf{LIBERO-Object}},
  yticklabels={,,}]
\addplot[fill=blue!65,draw=blue!80,error bars/.cd,y dir=both,y explicit]
  coordinates{(1,0.733)+-(0,0.262)(2,0.677)+-(0,0.216)};
\addplot[fill=red!65,draw=red!80,error bars/.cd,y dir=both,y explicit]
  coordinates{(1,0.663)+-(0,0.243)(2,0.580)+-(0,0.267)};
\end{axis}
\end{tikzpicture}
\end{subfigure}
\begin{subfigure}[b]{0.32\linewidth}
\centering
\begin{tikzpicture}
\begin{axis}[RL4ILbar,
  title={\small\textbf{LIBERO-Spatial}},
  yticklabels={,,}]
\addplot[fill=blue!65,draw=blue!80,error bars/.cd,y dir=both,y explicit]
  coordinates{(1,0.540)+-(0,0.200)(2,0.441)+-(0,0.198)};
\addplot[fill=red!65,draw=red!80,error bars/.cd,y dir=both,y explicit]
  coordinates{(1,0.455)+-(0,0.164)(2,0.429)+-(0,0.259)};
\end{axis}
\end{tikzpicture}
\end{subfigure}
\caption{Ablation study: soft fusion vs.\ hard argmax selection.}
\label{Figure:ablation-fusion}
\end{figure}
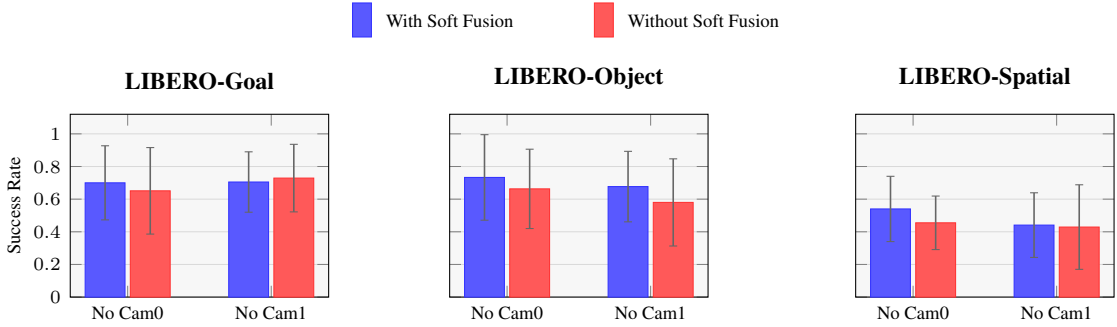

\paragraph{Effect of soft fusion.}
\cref{Figure:ablation-fusion} compares the full RL4IL pipeline, which
uses a soft cross-attention fusion head to aggregate signals from the
top-ranked candidate demonstrations, against a hard-argmax variant that
commits to a single retrieved demonstration without any aggregation.
Soft fusion consistently matches or outperforms the hard-argmax variant
across all three benchmark suites and both camera-dropout conditions,
demonstrating that attending over a pool of top-ranked candidates produces
more reliable action selection than selecting a single best match.
This is particularly evident under challenging dropout conditions where
individual retrieved demonstrations may be imperfect, and aggregation
over multiple candidates compensates for retrieval noise.

% ================================================================
%  FIGURE 1 — Soft Fusion vs No Soft Fusion  (SHORT CAPTION)
% ================================================================

% ================================================================
%  IN-TEXT EXPLANATION — paste before Figure 2 in your results section
% ================================================================

\begin{figure}[t]
\centering
\begin{center}
\begin{tikzpicture}
  \draw[fill=blue!65, draw=blue!80] (0,0) rectangle (0.25,0.45);
  \node[anchor=west, font=\scriptsize] at (0.32,0.22) {With Normalisation};
  \draw[fill=red!65, draw=red!80] (3.2,0) rectangle (3.45,0.45);
  \node[anchor=west, font=\scriptsize] at (3.52,0.22) {Without Normalisation};
\end{tikzpicture}
\end{center}
\vspace{2pt}
\begin{subfigure}[b]{0.32\linewidth}
\centering
\begin{tikzpicture}
\begin{axis}[RL4ILbar,
  title={\small\textbf{LIBERO-Goal}},
  ylabel={\scriptsize Success Rate},
  ylabel style={yshift=-2pt}]
\addplot[fill=blue!65,draw=blue!80,error bars/.cd,y dir=both,y explicit]
  coordinates{(1,0.700)+-(0,0.227)(2,0.705)+-(0,0.185)};
\addplot[fill=red!65,draw=red!80,error bars/.cd,y dir=both,y explicit]
  coordinates{(1,0.809)+-(0,0.216)(2,0.711)+-(0,0.155)};
\end{axis}
\end{tikzpicture}
\end{subfigure}
\begin{subfigure}[b]{0.32\linewidth}
\centering
\begin{tikzpicture}
\begin{axis}[RL4ILbar,
  title={\small\textbf{LIBERO-Object}},
  yticklabels={,,}]
\addplot[fill=blue!65,draw=blue!80,error bars/.cd,y dir=both,y explicit]
  coordinates{(1,0.733)+-(0,0.262)(2,0.677)+-(0,0.216)};
\addplot[fill=red!65,draw=red!80,error bars/.cd,y dir=both,y explicit]
  coordinates{(1,0.500)+-(0,0.371)(2,0.755)+-(0,0.151)};
\end{axis}
\end{tikzpicture}
\end{subfigure}
\begin{subfigure}[b]{0.32\linewidth}
\centering
\begin{tikzpicture}
\begin{axis}[RL4ILbar,
  title={\small\textbf{LIBERO-Spatial}},
  yticklabels={,,}]
\addplot[fill=blue!65,draw=blue!80,error bars/.cd,y dir=both,y explicit]
  coordinates{(1,0.540)+-(0,0.200)(2,0.441)+-(0,0.198)};
\addplot[fill=red!65,draw=red!80,error bars/.cd,y dir=both,y explicit]
  coordinates{(1,0.489)+-(0,0.246)(2,0.492)+-(0,0.221)};
\end{axis}
\end{tikzpicture}
\end{subfigure}
\caption{Ablation study: modality-fair normalisation vs.\ unnormalised embeddings.}
\label{Figure:ablation-norm}
\end{figure}
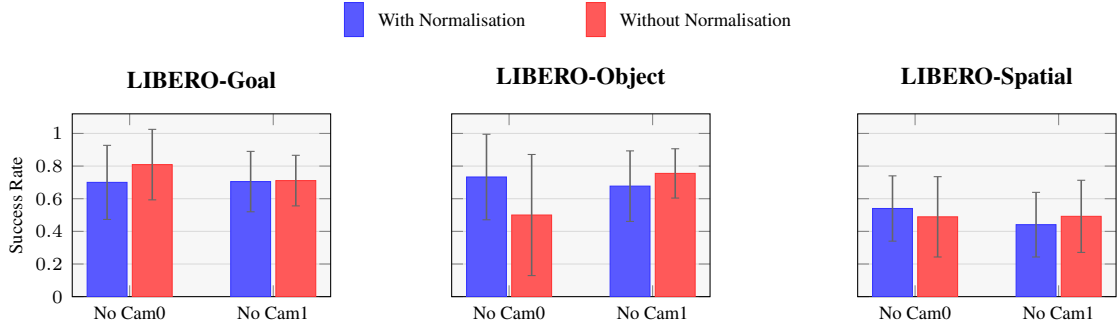

\paragraph{Effect of modality-fair normalisation.}
\cref{Figure:ablation-norm} compares RL4IL with and without
modality-fair $z$-score normalisation and dimension scaling.
Although the unnormalised variant occasionally achieves higher scores
on individual conditions, this is an artefact rather than a genuine
improvement: without normalisation, the modality with the largest
embedding norm dominates the distance metric, which can accidentally
favour retrieval when that single modality happens to be sufficient
for the task at hand.
This dominance, however, is dataset-dependent and does not generalise
reliably.
Normalisation removes this bias by ensuring all modalities contribute
equally to the retrieval distance regardless of their dimensionality,
producing more consistent performance across all suites and dropout
conditions and making the method behaviour more predictable when sensors
fail.

% ================================================================
%  FIGURE 2 — With Normalisation vs Without  (SHORT CAPTION)
% ================================================================

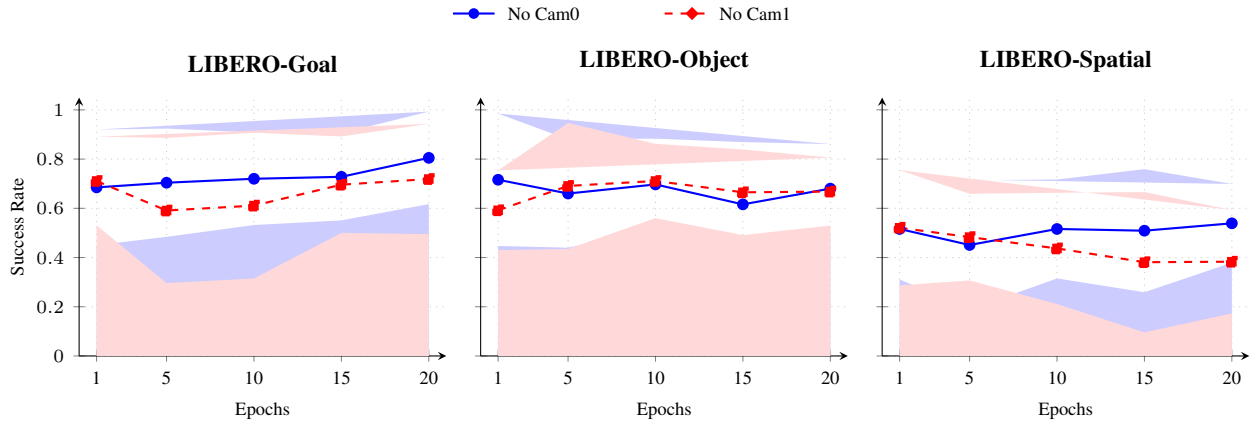
\begin{figure}[t]
\centering

% --- shared legend (once only) ---
\begin{tikzpicture}
  \draw[blue,  thick]         (0,0.15) -- (0.55,0.15);
  \fill[blue]                 (0.275,0.15) circle (2pt);
  \node[anchor=west, font=\scriptsize] at (0.6,0.15) {No Cam0};
  \draw[red, thick, dashed]   (2.8,0.15) -- (3.35,0.15);
  \fill[red]                  (3.075,0.15) -- ++(2.5pt,2.5pt) -- ++(2.5pt,-2.5pt)
                              -- ++(-2.5pt,-2.5pt) -- cycle;
  \node[anchor=west, font=\scriptsize] at (3.4,0.15) {No Cam1};
\end{tikzpicture}

\vspace{4pt}

% --- LIBERO-Goal ---
\begin{tikzpicture}
\begin{axis}[epochline,
  title={\small\textbf{LIBERO-Goal}},
  ylabel={\scriptsize Success Rate},
  ylabel style={yshift=-2pt}]

\addplot[blue!20,fill=blue!20,draw=none,forget plot]
  coordinates{(1,0.920)(5,0.924)(10,0.908)(15,0.905)(20,0.993)};
\addplot[blue!20,fill=blue!20,draw=none,forget plot]
  coordinates{(1,0.450)(5,0.484)(10,0.532)(15,0.551)(20,0.617)}\closedcycle;
\addplot[red!15,fill=red!15,draw=none,forget plot]
  coordinates{(1,0.891)(5,0.886)(10,0.907)(15,0.892)(20,0.943)};
\addplot[red!15,fill=red!15,draw=none,forget plot]
  coordinates{(1,0.531)(5,0.296)(10,0.315)(15,0.500)(20,0.495)}\closedcycle;

\addplot[blue,thick,mark=*,mark size=1.8pt]
  coordinates{(1,0.685)(5,0.704)(10,0.720)(15,0.728)(20,0.805)};
\addplot[red,thick,dashed,mark=square*,mark size=1.8pt]
  coordinates{(1,0.711)(5,0.591)(10,0.611)(15,0.696)(20,0.719)};
\end{axis}
\end{tikzpicture}
\hfill
% --- LIBERO-Object ---
\begin{tikzpicture}
\begin{axis}[epochline noy,
  title={\small\textbf{LIBERO-Object}}]

\addplot[blue!20,fill=blue!20,draw=none,forget plot]
  coordinates{(1,0.985)(5,0.880)(10,0.883)(15,0.869)(20,0.861)};
\addplot[blue!20,fill=blue!20,draw=none,forget plot]
  coordinates{(1,0.447)(5,0.440)(10,0.511)(15,0.363)(20,0.499)}\closedcycle;
\addplot[red!15,fill=red!15,draw=none,forget plot]
  coordinates{(1,0.754)(5,0.947)(10,0.862)(15,0.839)(20,0.806)};
\addplot[red!15,fill=red!15,draw=none,forget plot]
  coordinates{(1,0.430)(5,0.435)(10,0.560)(15,0.491)(20,0.530)}\closedcycle;

\addplot[blue,thick,mark=*,mark size=1.8pt]
  coordinates{(1,0.716)(5,0.660)(10,0.697)(15,0.616)(20,0.680)};
\addplot[red,thick,dashed,mark=square*,mark size=1.8pt]
  coordinates{(1,0.592)(5,0.691)(10,0.711)(15,0.665)(20,0.668)};
\end{axis}
\end{tikzpicture}
\hfill
% --- LIBERO-Spatial ---
\begin{tikzpicture}
\begin{axis}[epochline noy,
  title={\small\textbf{LIBERO-Spatial}}]

\addplot[blue!20,fill=blue!20,draw=none,forget plot]
  coordinates{(1,0.721)(5,0.714)(10,0.717)(15,0.759)(20,0.699)};
\addplot[blue!20,fill=blue!20,draw=none,forget plot]
  coordinates{(1,0.311)(5,0.188)(10,0.315)(15,0.259)(20,0.379)}\closedcycle;
\addplot[red!15,fill=red!15,draw=none,forget plot]
  coordinates{(1,0.755)(5,0.659)(10,0.663)(15,0.666)(20,0.593)};
\addplot[red!15,fill=red!15,draw=none,forget plot]
  coordinates{(1,0.287)(5,0.307)(10,0.211)(15,0.096)(20,0.173)}\closedcycle;

\addplot[blue,thick,mark=*,mark size=1.8pt]
  coordinates{(1,0.516)(5,0.451)(10,0.516)(15,0.509)(20,0.539)};
\addplot[red,thick,dashed,mark=square*,mark size=1.8pt]
  coordinates{(1,0.521)(5,0.483)(10,0.437)(15,0.381)(20,0.383)};
\end{axis}
\end{tikzpicture}

\caption{Effect of training epochs on task success rate under camera dropout.
Solid blue = agent-view absent (No Cam0); dashed red = in-hand absent (No Cam1).
Shaded regions denote $\pm1$ standard deviation.}
\label{fig:epoch-study}
\end{figure}

\paragraph{Effect of training epochs.} We investigate the effect of training epochs $\{1, 5, 10, 15, 20\}$ on task success rate under camera dropout across all three LIBERO suites. As shown in \cref{fig:epoch-study}, performance on LIBERO-Spatial and LIBERO-Object remains relatively stable across epochs. In contrast, LIBERO-Goal exhibits a clear upward trend with longer training, demonstrating that additional training epochs yield meaningful gains under goal-conditioned tasks. Notably, RL4IL maintains competitive performance even at epoch 1 across all suites, which surpasses the reported results of all baselines in \cref{table:dropout-comparison}.

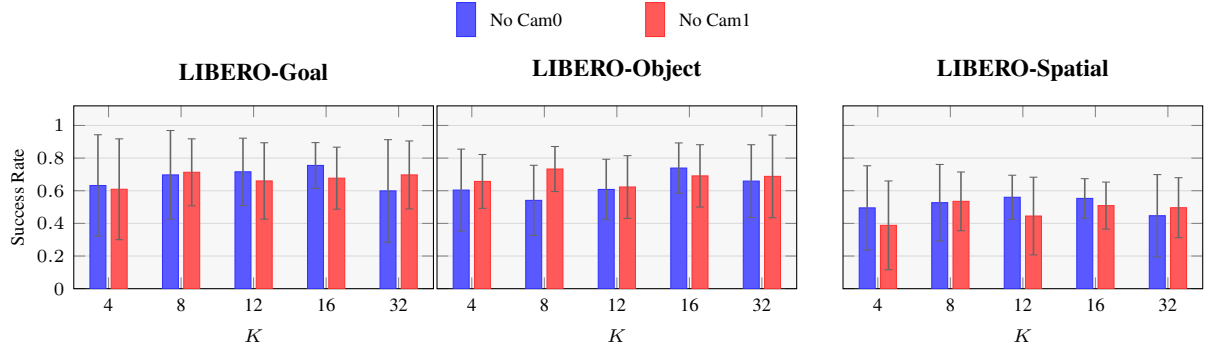
\begin{figure}[t]
\centering

% ---- legend (once) ----
\begin{tikzpicture}
  \draw[fill=blue!65, draw=blue!80] (0,0) rectangle (0.25,0.45);
  \node[anchor=west, font=\scriptsize] at (0.32,0.22) {No Cam0};
  \draw[fill=red!65,  draw=red!80]  (2.5,0) rectangle (2.75,0.45);
  \node[anchor=west, font=\scriptsize] at (2.82,0.22) {No Cam1};
\end{tikzpicture}

\vspace{4pt}

\begin{subfigure}[b]{0.32\linewidth}
\centering
\begin{tikzpicture}
\begin{axis}[RL4ILtopk,
  title={\small\textbf{LIBERO-Goal}},
  ylabel={\scriptsize Success Rate},
  ylabel style={yshift=-2pt}]
\addplot[fill=blue!65,draw=blue!80,error bars/.cd,y dir=both,y explicit]
  coordinates{(1,0.632)+-(0,0.311)(2,0.697)+-(0,0.272)(3,0.716)+-(0,0.206)(4,0.755)+-(0,0.140)(5,0.599)+-(0,0.314)};
\addplot[fill=red!65,draw=red!80,error bars/.cd,y dir=both,y explicit]
  coordinates{(1,0.609)+-(0,0.309)(2,0.713)+-(0,0.205)(3,0.660)+-(0,0.234)(4,0.677)+-(0,0.190)(5,0.697)+-(0,0.208)};
\end{axis}
\end{tikzpicture}
\end{subfigure}
\begin{subfigure}[b]{0.32\linewidth}
\centering
\begin{tikzpicture}
\begin{axis}[RL4ILtopk,
  title={\small\textbf{LIBERO-Object}},
  yticklabels={,,}]
\addplot[fill=blue!65,draw=blue!80,error bars/.cd,y dir=both,y explicit]
  coordinates{(1,0.604)+-(0,0.251)(2,0.541)+-(0,0.215)(3,0.608)+-(0,0.185)(4,0.739)+-(0,0.154)(5,0.659)+-(0,0.223)};
\addplot[fill=red!65,draw=red!80,error bars/.cd,y dir=both,y explicit]
  coordinates{(1,0.657)+-(0,0.165)(2,0.733)+-(0,0.138)(3,0.623)+-(0,0.192)(4,0.691)+-(0,0.191)(5,0.688)+-(0,0.253)};
\end{axis}
\end{tikzpicture}
\end{subfigure}
\begin{subfigure}[b]{0.32\linewidth}
\centering
\begin{tikzpicture}
\begin{axis}[RL4ILtopk,
  title={\small\textbf{LIBERO-Spatial}},
  yticklabels={,,}]
\addplot[fill=blue!65,draw=blue!80,error bars/.cd,y dir=both,y explicit]
  coordinates{(1,0.495)+-(0,0.258)(2,0.527)+-(0,0.234)(3,0.560)+-(0,0.135)(4,0.553)+-(0,0.121)(5,0.447)+-(0,0.252)};
\addplot[fill=red!65,draw=red!80,error bars/.cd,y dir=both,y explicit]
  coordinates{(1,0.388)+-(0,0.272)(2,0.535)+-(0,0.180)(3,0.445)+-(0,0.238)(4,0.509)+-(0,0.144)(5,0.496)+-(0,0.184)};
\end{axis}
\end{tikzpicture}
\end{subfigure}

\caption{Ablation study: effect of top-$K$ fusion candidates
($K' = K_{\mathrm{imp}}' \in \{4,8,12,16,32\}$) on task success rate.
Blue = agent-view camera absent (No Cam0); red = in-hand camera absent (No Cam1).}
\label{fig:topk-ablation}
\end{figure}

\paragraph{Effect of top-$K$ fusion candidates.} We investigate the effect of the number of top-$K$ fusion candidates
($K' = K_{\mathrm{imp}}' \in \{4, 8, 12, 16, 32\}$) on task success
rate under camera dropout, and results shown in
\cref{fig:topk-ablation}.
Across all three suites, performance is generally stable for moderate
values of $K$, suggesting that RL4IL is not highly sensitive to this
hyperparameter within a reasonable range.

\section{Conclusion}
\label{sec:conclusion}
    We presented RL4IL, a reinforcement learning-guided retrieval framework
    for imitation learning that is robust to sensor dropout without requiring
    any policy retraining.
    At its core, a PPO-trained policy operates over BFS-augmented candidate
    sets to select the most relevant expert demonstration from a frozen
    training library --- the first application of reinforcement learning to
    nearest-neighbour selection in imitation learning.
    A soft cross-attention fusion head aggregates signals from the top-ranked
    candidates, consistently outperforming hard argmax selection across all
    evaluated conditions.
    When a modality is absent at inference time, a dedicated per-modality RL
    retrieval policy identifies donor demonstrations and a soft imputation
    head reconstructs the missing embedding via cross-attention, restoring the
    full observation representation without any modification to the downstream
    pipeline.
    
    Experiments on three LIBERO benchmark suites --- LIBERO-Spatial,
    LIBERO-Object, and LIBERO-Goal --- demonstrate that RL4IL substantially
    outperforms all baselines under complete camera dropout, achieving success
    rates of up to $0.733$ where the strongest prior method, DisDP, reaches
    only $0.295$.
    Ablation studies confirm that soft fusion, modality-fair normalisation,
    and the RL retrieval policy each contribute meaningfully to overall
    performance, and that RL4IL maintains competitive results even with a
    single training epoch.
    
    These results suggest that retrieval-based imitation learning, augmented
    with learned ranking and soft imputation, is a practical and effective
    strategy for deploying robotic systems in real-world environments where
    sensor availability cannot be guaranteed.
    Future work may explore extending RL4IL to online settings, incorporating
    richer temporal context into the retrieval query, and scaling the
    demonstration library to larger and more diverse task distributions.

%===============================================================================

\section*{acknowledgments}
This research work has received funding from the EU through the ARTEMIS project (grant agreement No 101136299) and PRESERVE project (grant agreement No 101168309). Authors confirm that they did experiments on GPU resources at Bournemouth University, and also there is no conflict of interests.

%===============================================================================

\bibliographystyle{unsrt}  
\bibliography{references}

\newpage
\begin{appendix}
% -----------------------------------------------------------
\section{Modality-Specific Normalisation and Modality-Fair Distance}
\label{app:norm-dist}

    Raw embeddings produced by heterogeneous encoders can differ dramatically in
    scale and per-dimension variance, which would cause the distance measure
    introduced below to be dominated by modalities with larger raw magnitudes.
    To place all modalities on a common footing we fit modality-specific z-score
    statistics on the training set.
    For each modality $m$ we compute the element-wise mean and standard deviation
    over all $N$ training samples:
    
    \begin{equation}
        \boldsymbol{\mu}_m
            = \frac{1}{N}\sum_{i=1}^{N}\mathbf{z}_i^{(m)},
        \qquad
        \boldsymbol{\sigma}_m
            = \sqrt{\frac{1}{N-1}
              \sum_{i=1}^{N}
              \bigl(\mathbf{z}_i^{(m)}-\boldsymbol{\mu}_m\bigr)^{\!2}},
        \label{eq:norm-stats}
    \end{equation}
    
    \noindent where all operations are element-wise.
    Any dimension whose standard deviation falls below $\epsilon>0$ is clamped to
    $1$, preventing division by zero while leaving near-constant dimensions with
    negligible influence on distances.
    Normalised embeddings are then obtained via
    
    \begin{equation}
        \tilde{\mathbf{z}}_i^{(m)}
            = \frac{\mathbf{z}_i^{(m)} - \boldsymbol{\mu}_m}
                   {\boldsymbol{\sigma}_m},
        \label{eq:norm}
    \end{equation}
    
    \noindent applied element-wise.
    At inference time, test embeddings are normalised with the training statistics
    $\boldsymbol{\mu}_m$, $\boldsymbol{\sigma}_m$, so no test information leaks
    into the normalisation.

    Given two normalised multimodal samples
    $\tilde{\mathbf{z}}_i = \{\tilde{\mathbf{z}}_i^{(m)}\}_{m=1}^{M}$ and
    $\tilde{\mathbf{z}}_j = \{\tilde{\mathbf{z}}_j^{(m)}\}_{m=1}^{M}$,
    we define their dissimilarity as
    
    \begin{equation}
        d(i,j)
            = \sqrt{
                \frac{1}{M}
                \sum_{m=1}^{M}
                \frac{\bigl\|\tilde{\mathbf{z}}_i^{(m)}
                            -\tilde{\mathbf{z}}_j^{(m)}\bigr\|_2^2}
                     {d_m}
              },
        \label{eq:dist}
    \end{equation}
    
    \noindent where $d_m$ is the embedding dimensionality of modality $m$.
    The inner division by $d_m$ converts each modality's squared $\ell_2$ error
    into a mean squared error per dimension, so that a modality with a larger
    embedding space does not disproportionately dominate the distance.
    The outer factor $1/M$ averages the per-modality contributions so that the
    distance is comparable across pairs regardless of how many modalities are
    present.

\section{State and Candidate Features}
\label{app:method:features}

The state feature vector $\phi(\mathbf{s})\in\mathbb{R}^{d+2}$ for source
$\mathbf{s}$ is

\begin{equation}
    \phi(\mathbf{s})
        = \bigl[\mathbf{f}_{\mathbf{s}};\;
                \mathrm{Var}(y_{\mathcal{B}(\mathbf{s})});\;
                |\mathcal{B}(\mathbf{s})|\bigr],
    \label{eq:state}
\end{equation}

\noindent where $\mathbf{f}_{\mathbf{s}}\in\mathbb{R}^d$ is the normalised
concatenated embedding of $\mathbf{s}$,
$\mathrm{Var}(y_{\mathcal{B}(\mathbf{s})})$ is the variance of the labels of
all nodes in the BFS set, and $|\mathcal{B}(\mathbf{s})|$ is the set size.

The candidate feature vector
$\psi(\mathbf{s},\mathbf{v})\in\mathbb{R}^{d+4}$ for
$\mathbf{v}\in\mathcal{B}(\mathbf{s})$ is

\begin{equation}
    \psi(\mathbf{s},\mathbf{v})
        = \left[\mathbf{f}_{\mathbf{v}};\;
                \frac{\delta(\mathbf{s},\mathbf{v})}
                     {\max_{\mathbf{u}\in\mathcal{B}}\delta(\mathbf{s},\mathbf{u})};\;
                \frac{\mathrm{depth}(\mathbf{s},\mathbf{v})}
                     {\max_{\mathbf{u}\in\mathcal{B}}\mathrm{depth}(\mathbf{s},\mathbf{u})};\;
                \frac{\mathrm{rank}_d(\mathbf{v})}
                     {|\mathcal{B}(\mathbf{s})|-1};\;
                y_{\mathbf{v}} - \bar{y}_{\mathcal{B}(\mathbf{s})}
          \right],
    \label{eq:cand-feat}
\end{equation}

\noindent where $\mathrm{depth}(\mathbf{s},\mathbf{v})$ is the number of hops
in the shortest path, $\mathrm{rank}_d(\mathbf{v})$ is the $0$-indexed rank
of $\mathbf{v}$ when $\mathcal{B}(\mathbf{s})$ is sorted by ascending graph
distance, and $\bar{y}_{\mathcal{B}(\mathbf{s})}$ is the mean label in the
set.
All normalised features lie in $[0,1]$.

\section{Test-Time Set Size Estimation}
\label{app:method:impl:test-size}

At test time the true label of a query is unknown, so the BFS cannot be
stopped at the oracle.
Instead we estimate the appropriate set size using the training set sizes of
the query's initial neighbours (UNN or $k$-NN seeds) weighted by their
inverse distances:

\begin{equation}
    \hat{S}(\mathbf{o})
        = \displaystyle\max_{\mathbf{n}\in\mathcal{C}(\mathbf{o})}|\mathcal{B}(\mathbf{n})|             
    \label{eq:test-size}
\end{equation}

\noindent where $|\mathcal{B}(\mathbf{n})|$ is the training BFS set size of
neighbour $\mathbf{n}$, pre-computed and stored during training.
The BFS from $\mathbf{o}$ is then expanded until exactly $\hat{S}(\mathbf{o})$
nodes have been collected, after which the policy selects the best candidate.
This ensures that the distribution of set sizes seen by the policy at test
time matches the distribution encountered during training.

\section{Hyperparameters}
\label{app:hyperparams}

All hyperparameters are shared across the three LIBERO benchmark suites
(LIBERO-Spatial, LIBERO-Object, and LIBERO-Goal) unless stated otherwise.
We group them by pipeline stage.

\subsection*{Encoders and Embeddings}

Each of the three modalities --- agent-view camera, in-hand camera, and
natural-language instruction --- is encoded by a frozen CLIP ViT-B/32
model, producing a $d_m = 512$-dimensional embedding per modality.
For visual modalities, $N_{\mathrm{frames}} = 8$ frames are sampled
uniformly from each demonstration and their CLIP image features are
averaged to form the per-demonstration visual embedding.
No encoder weights are updated at any stage.

\subsection*{Neighbourhood and Graph Construction}

The initial neighbour set used for BFS seeding is constructed with
$k_{\mathrm{approx}} = 20$ approximate nearest neighbours (used to
form the candidate pool) and $k_{\mathrm{seed}} = 5$ seed neighbours
passed to the BFS.
The $k$-NN graph over which BFS expands uses $k_{\mathrm{graph}} = 5$
edges per node.
BFS is bounded by a maximum depth of $D = 6$ hops and a maximum of
$200$ visited nodes per source, whichever is reached first.
The training set is split with a validation fraction of $0.15$; all
BFS statistics are computed on training embeddings only, and test
embeddings are normalised using training-set statistics.

\subsection*{Prediction RL Policy (PPO)}

The RL policy is trained with Proximal Policy Optimisation for
$30$ epochs using the clipped surrogate objective with
clipping threshold $\varepsilon = 0.2$.
The entropy bonus coefficient is set to
$\alpha_{\mathrm{ent}} = 10^{-4}$.
The Adam optimiser is used with learning rate $3 \times 10^{-4}$
and gradient norm clipping at $1.0$.
Rollout minibatches contain $128$ samples.

\subsection*{Soft Prediction Fusion Head}

The fusion head attends over the top-$K' = 32$ RL-ranked candidate
demonstrations.
Query and key projections map embeddings into a shared space of
dimension $d_f = 128$, split across $H = 4$ attention heads.
The head is trained for $30$ epochs with the Adam optimiser
at learning rate $3 \times 10^{-4}$, a batch size of $32$,
and gradient norm clipping at $1.0$.

\subsection*{Imputation RL Policy (PPO)}

A dedicated per-modality imputation policy is trained independently
for each camera modality.
Training runs for $50$ PPO epochs with minibatches of $64$ samples,
entropy coefficient $\alpha_{\mathrm{ent}} = 10^{-4}$, learning
rate $3 \times 10^{-4}$, and gradient norm clipping at $1.0$.
The clipping threshold and optimiser are identical to the prediction
RL policy.
The language modality is always present and is never subject to
imputation.

\subsection*{Soft Imputation Head}

The soft imputation head aggregates the top-$K_{\mathrm{imp}}' = 32$
RL-ranked donor embeddings via cross-attention.
Projections map into a shared space of dimension
$d_{\mathrm{imp}} = 64$ across $H_{\mathrm{imp}} = 2$ attention heads.
The head is trained for $30$ epochs under a supervised MSE objective
with the Adam optimiser at learning rate $3 \times 10^{-4}$,
batch size $32$, and gradient norm clipping at $1.0$.
The imputation RL policy is frozen throughout.

\subsection*{Rollout Evaluation Protocol}

Task success is evaluated across $3$ random seeds, with $25$ rollouts
per task per seed, for a total of $75$ rollouts per task.
Each rollout is allowed a maximum of $260$ environment steps.
Success rate is reported as the fraction of rollouts in which the task
is completed within the step budget, averaged over seeds, following
the evaluation protocol of DisDP~\citep{vanjani2025disdp}.

\end{appendix}
\end{document}